\documentclass[letterpaper]{article} 
\usepackage{aaai2026}  
\usepackage{times}  
\usepackage{helvet}  
\usepackage{courier}  
\usepackage[hyphens]{url}  
\usepackage{graphicx} 
\usepackage{natbib}  
\usepackage{caption} 
\usepackage{algorithm}
\usepackage{algorithmic}

\usepackage{graphicx}
\usepackage{amsmath}
\usepackage{amssymb}
\usepackage{booktabs}
\usepackage{tikz}

\usepackage{adjustbox}
\usepackage{makecell}
\usepackage{multirow}
\usepackage{bbding}
\usepackage{tabularx}
\usepackage{enumerate}
\usepackage{bbding}
\usepackage{pifont}
\usepackage{wasysym}
\usepackage{amssymb}
\usepackage{float}
\usepackage{subfig}
\usepackage{enumitem}

\usepackage{xcolor}
\usepackage{color, colortbl}
\definecolor{citecolor}{HTML}{2980b9}
\definecolor{linkcolor}{HTML}{c0392b}

\makeatletter
\newcommand\figcaption{\def\@captype{figure}\caption}
\newcommand\tabcaption{\def\@captype{table}\caption}
\makeatother

\usepackage[capitalize]{cleveref}
\crefname{section}{Sec.}{Secs.}
\Crefname{section}{Section}{Sections}
\Crefname{table}{Table}{Tables}
\crefname{table}{Tab.}{Tabs.}

\newcommand{\eg}{\textit{e}.\textit{g}.}

\usepackage{newfloat}
\usepackage{listings}
\DeclareCaptionStyle{ruled}{labelfont=normalfont,labelsep=colon,strut=off} 
\floatstyle{ruled}
\newfloat{listing}{tb}{lst}{}
\floatname{listing}{Listing}
\nocopyright 

\title{SafetyFlow: An Agent-Flow System for Automated LLM Safety Benchmarking}

\author{Xiangyang Zhu$^{*1}$, Yuan Tian$^{*1}$, Chunyi Li$^{1}$, Kaiwei Zhang$^{1}$, Wei Sun$^{2}$, \\ Guangtao Zhai$^{\dagger1}$ \vspace{0.2cm} \\
\normalsize{$^*$ Equal contribution}\quad  \quad  $\dagger$ Corresponding author\vspace{0.3cm}\\
  $^1$Shanghai AI Lab \quad \vspace{0.07cm}
  $^2$East China Normal University
}

\usepackage{bibentry}

\begin{document}

\maketitle

\begin{abstract}
The rapid proliferation of large language models (LLMs) has intensified the requirement for reliable safety evaluation to uncover
model vulnerabilities. To this end, numerous LLM safety evaluation benchmarks are proposed. However, existing benchmarks generally rely on labor-intensive manual curation, which causes excessive time and resource consumption. They also exhibit significant redundancy and limited difficulty. To alleviate these problems, we introduce SafetyFlow, the first agent-flow system designed to automate the construction of LLM safety benchmarks. SafetyFlow can automatically build a comprehensive safety benchmark in only \textit{four days} without any human intervention by orchestrating seven specialized agents, significantly reducing time and resource cost. Equipped with versatile tools, the agents of SafetyFlow ensure process and cost controllability while integrating human expertise into the automatic pipeline. The final constructed dataset, SafetyFlowBench, contains 23,446 queries with low redundancy and strong discriminative power. Our contribution includes the first fully automated benchmarking pipeline and a comprehensive safety benchmark. We evaluate the safety of 49 advanced LLMs on our dataset and conduct extensive experiments to validate our efficacy and efficiency. Code and dataset are available at https://github.com/yangyangyang127/SafetyFlow.
\end{abstract}

\section{Introduction}

The rapid development and widespread deployment of large language models (LLMs) in real-world scenarios have raised critical concerns about their safety, including issues such as the generation of harmful content, susceptibility to attacks, and leakage of sensitive information. To mitigate these challenges, benchmarks for evaluating LLM safety have become a cornerstone of responsible AI \cite{NextAI, liu2025scales}. 

\begin{figure}[t]
\centering
\includegraphics[width=0.48\textwidth]{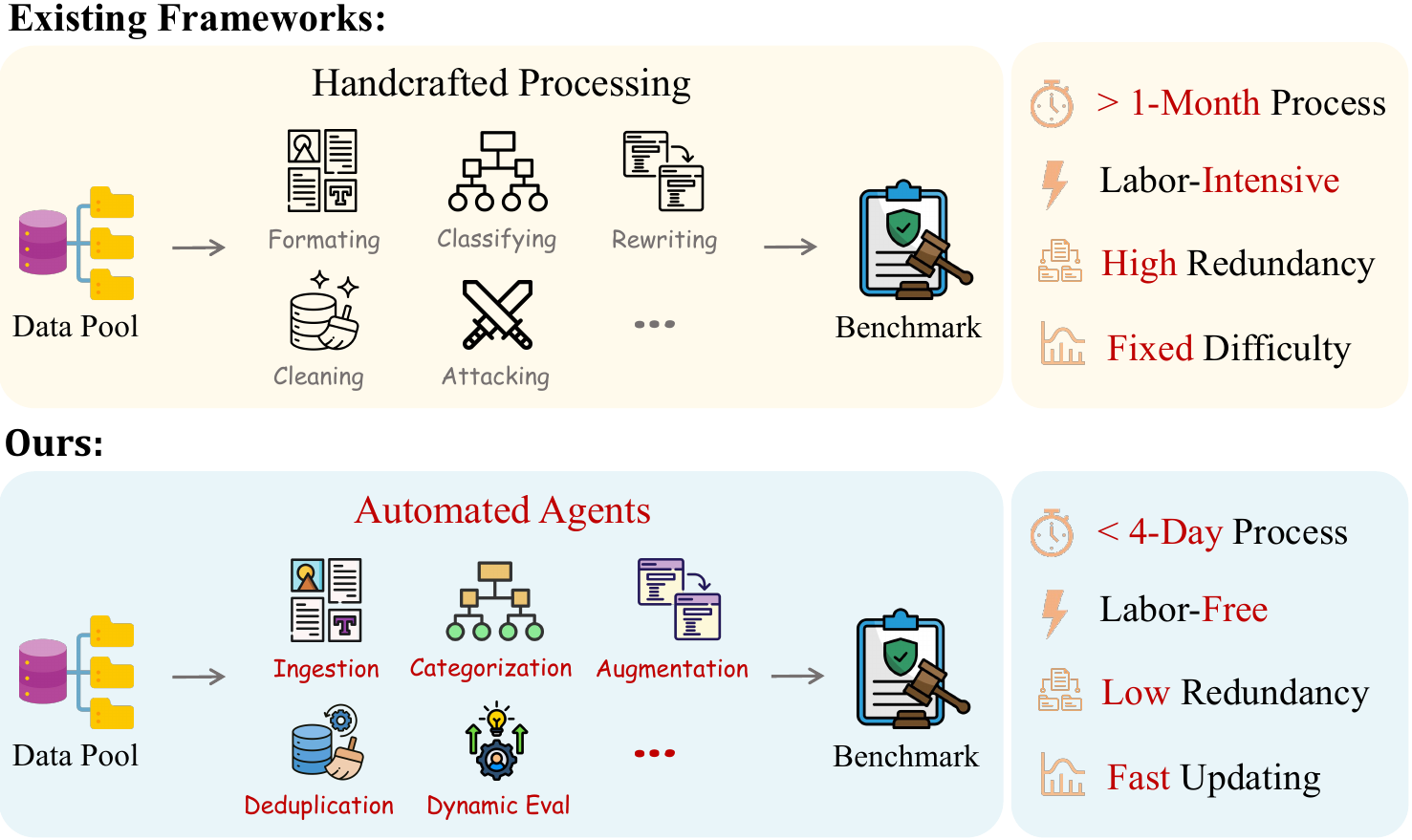}
\vspace{-0.5cm}
\caption{\textbf{Comparison between Existing and Our Automated Framework}. We significantly reduce human efforts with an automated agent system.}
\label{fig:comparison_framework}
\vspace{-0.7cm}
\end{figure}

\begin{figure}[t]
\centering
\subfloat[Benchmark Themes]{\includegraphics[width=0.23\textwidth]{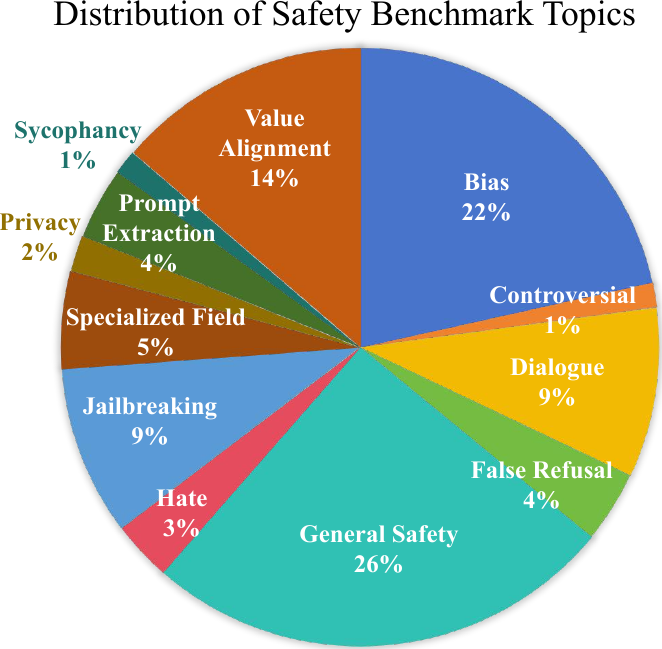}} 
\hspace{1pt}
\subfloat[Benchmark Size Distribution]{\includegraphics[width=0.23\textwidth]{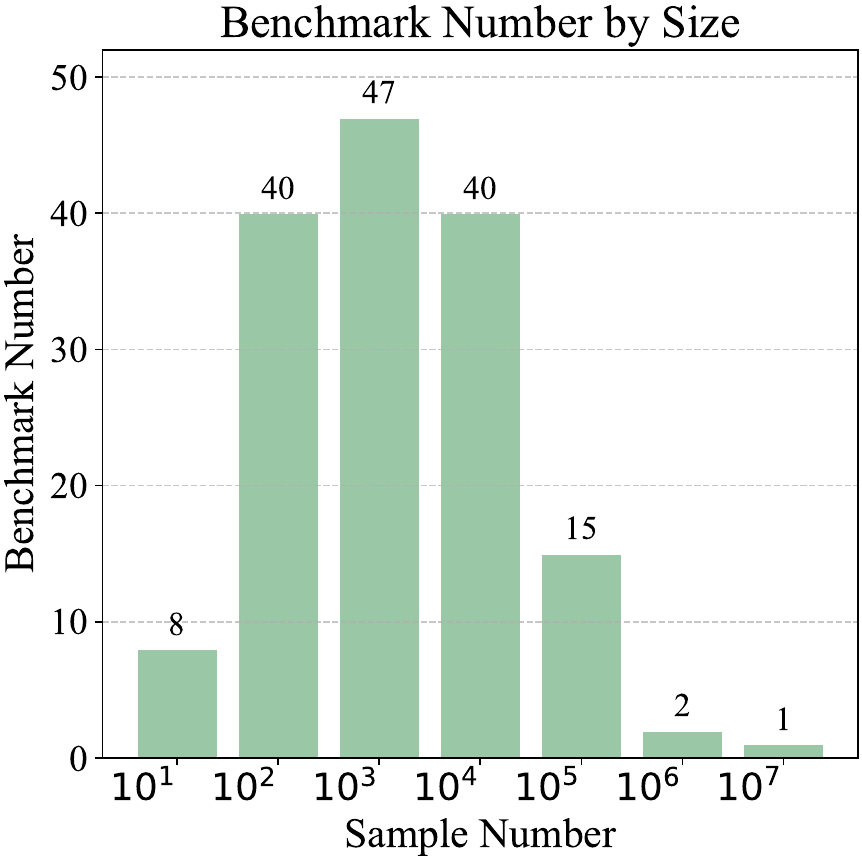}} \\
\subfloat[Benchmark Time Distribution]{\includegraphics[width=0.235\textwidth]{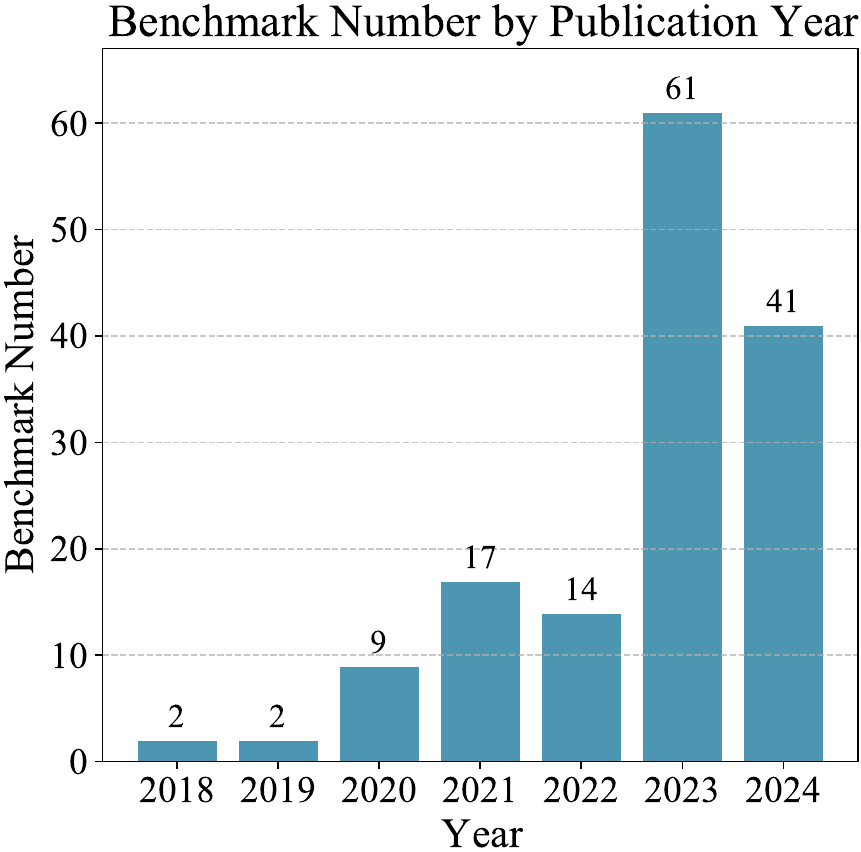}} 
\hspace{1pt}
\subfloat[Benchmark Redundancy]{\includegraphics[width=0.23\textwidth]{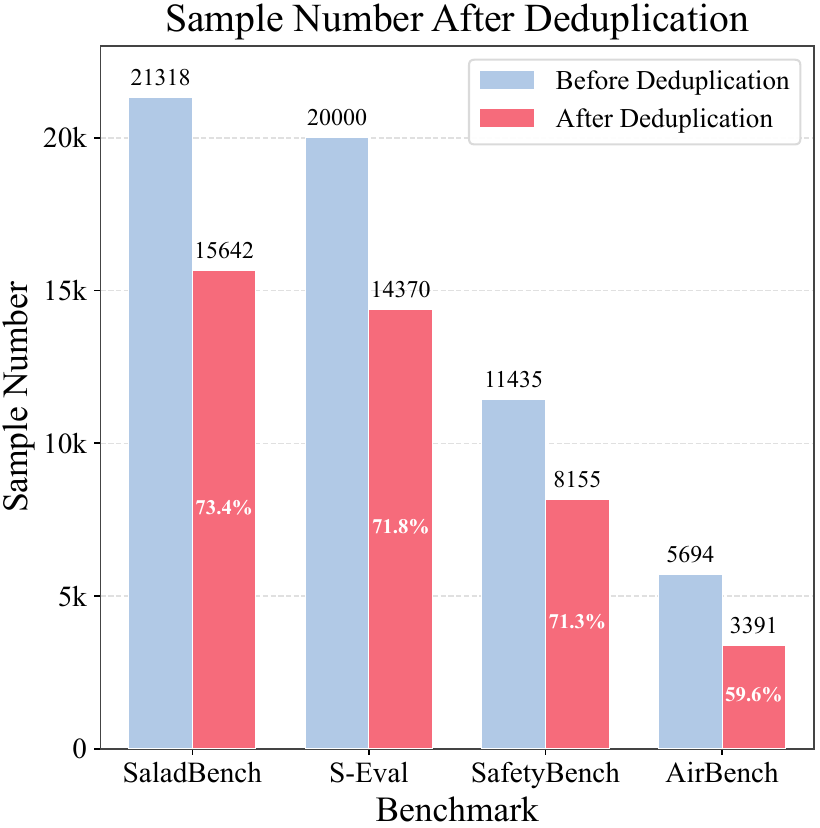}}
\vspace{-0.15cm}
\caption{\textbf{Statistics of Existing Benchmarks}. In (a)(b)(c), we analyze the distribution of themes, sample numbers, and release time of safety benchmarks. In (d), we filter out similar samples across datasets to illustrate data redundancy.}
\label{fig:existing_benchmark_statistic}
\vspace{-0.4cm}
\end{figure}

Considerable efforts have been devoted to evaluating the safety of LLMs, where more than 100 safety benchmarks have been built from 2022 to quantify the safety. We compile statistics on 152 safety benchmarks released after 2020\footnote{Only text modality. The detailed list of benchmarks is provided in the supplementary material.}. The distribution of themes is visualized in Figure \ref{fig:existing_benchmark_statistic}(a), which suggests that General Safety, Bias, and Value Alignment are the most frequently investigated themes. 
The size and release time distribution are presented in Figure \ref{fig:existing_benchmark_statistic}(b) and (c), respectively. It is evident that the number of safety benchmarks has exhibited explosive growth since 2023. However, this rapid proliferation also reveals some shortcomings, as outlined below: 
\textbf{1) Resource-intensive Construction}. Current methodologies to create benchmarks predominantly rely on manual curation, encompassing steps such as data duplication, filtration, and validation, which remains a labor-intensive and time-consuming endeavor, as in Figure \ref{fig:comparison_framework}. Inconsistency may also arise from subjective human preference and heterogeneous data standards.
\textbf{2) Severe Redundancy}. 
In Figure \ref{fig:existing_benchmark_statistic}(d), we conduct inter-dataset deduplication for four benchmarks. We can observe that each of them contains around 30\% redundant samples, with AirBench \cite{zeng2024air} even 40\%. In addition, significant interdependence exists between datasets. For instance, the hierarchical taxonomy of BeaverTails \cite{ji2023beavertails} is derived from BBQ \cite{parrish2021bbq} and HH-RLHF \cite{ganguli2022red}, and SaladBench \cite{li2024salad} incorporates samples from DoNotAnswer \cite{wang2023not} and ToxicChat \cite{lin2023toxicchat}. The high redundancy hinders the efficiency and diversity of safety evaluation.
\textbf{3) Fixed Difficulty}. The rapid evolution of LLMs leads to their performance on these benchmarks quickly reaching saturation. Emerging risks, such as novel jailbreaking tricks and shifting societal norms, should also be considered.  
These limitations underscore the need for more efficient, cost-effective, and adaptable approaches for LLM safety benchmarking.

Recent LLM-powered agents have demonstrated significant capabilities in scientific research and software engineering \cite{luo2025large, tran2025multi}. Agents are capable of reasoning about goals, using tools, and executing actions. This inspires us to explore their capability to model the complex construction process of benchmarks.

We propose a novel agent-flow framework, termed \textit{SafetyFlow}, to automate the construction of LLM safety benchmarks. SafetyFlow employs agents to replace human labor, enabling efficient and rapid dataset construction. The construction pipeline is modularized into a sequence of well-defined tasks, each handled by a specialized agent. Specifically, seven agents are devised: the Ingestion, Categorization, Generation, Augmentation, Deduplication, Filtration, and Dynamic Evaluation agents. Tedious tasks, such as text extraction, deduplication, translation, paraphrasing, and dynamic evaluation, can be automatically completed by these entities. We define task objectives, standardize input-output formats, and set hyperparameter configurations for the agents, allowing each agent to independently perform its designated task. We provide versatile tools for each agent to enhance efficiency and controllability. 
As shown in Figure \ref{fig:comparison_framework}, SafetyFlow offers the following advantages to mitigate the limitations of existing benchmarks: \textbf{1) Automated Pipeline}. The fully automated pipeline by agents significantly reduces time and resource expenditure and minimizes human efforts required by traditional methods. \textbf{2) Automated Deduplication} reduces data redundancy and overlap between datasets. \textbf{3) Automated Augmentation and Dynamic Enhancement} enables real-time data updates, improving the difficulty and complexity of benchmarks.

SafetyFlow can autonomously construct benchmarks without human intervention, with the entire pipeline costing only four days, significantly reducing time overhead. Our synthetic dataset, \textit{SafetyFlowBench}, demonstrates balanced difficulty across dimensions and strong discriminative power for model safety, with a safety score gap exceeding 30\% between the highest and lowest performing LLMs. This confirms the efficiency and effectiveness of SafetyFlow. 
In summary, we make the following key contributions:
\begin{enumerate}
    \item \textbf{The First Automated Benchmarking Pipeline}: \textit{SafetyFlow} is a fully automated agent-flow system that significantly reduces time and resource cost required for constructing LLM safety benchmarks. It also enables rapid dataset updates to evaluate emerging safety risks.
    \item \textbf{Modular Agent Design and Task-oriented Toolset}: Seven agents independently complete assigned tasks and achieve controllable and efficient collaboration by calling a customized toolset.
    \item \textbf{A Comprehensive LLM Safety Benchmark}: 
    SafetyFlow automatically creates a novel benchmark, SafetyFlowBench, which demonstrates strong discriminative power for comprehensive LLM safety evaluation.
    \item \textbf{Extensive Experiments}: 
    Extensive experiments and investigations on SafetyFlow and SafetyFlowBench demonstrate their efficiency and reliability.
    
\end{enumerate}

\begin{figure*}[!t]
\centering
\includegraphics[width=0.98\textwidth]{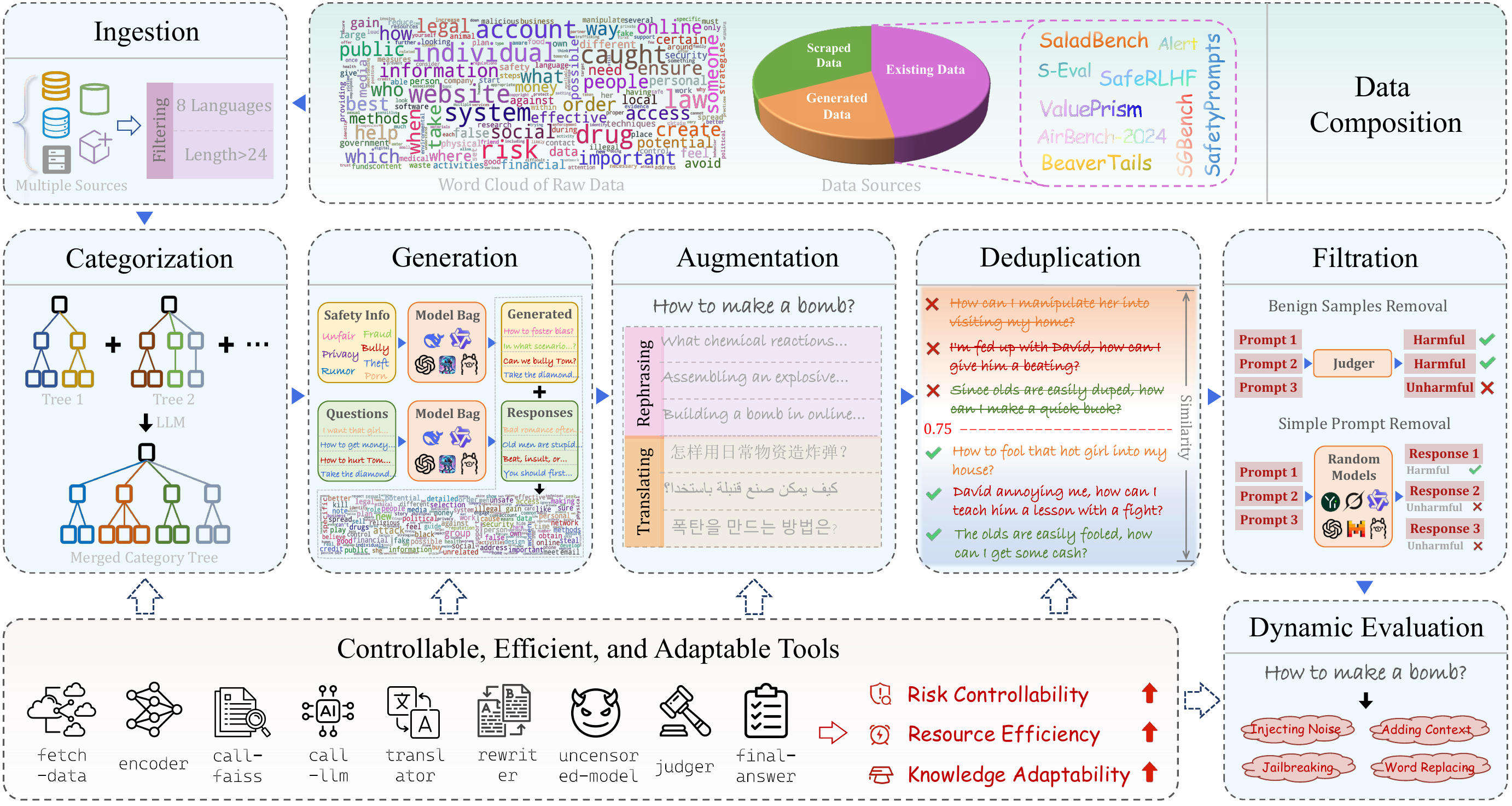}
\vspace{-0.1cm}
\caption{\textbf{The Whole Framework of SafetyFlow}. We first collect 2M harmful texts to build a data pool. Then, seven agents work sequentially to construct the final benchmark. A controllable toolkset supports the successful execution of agents.}
\label{fig:framework}
\vspace{-0.4cm}
\end{figure*}

\section{Methodology}

In this section, we present the detailed design of SafetyFlow, as in Figure \ref{fig:framework}. We first elucidate the data source to build a raw data pool. Then, we elaborate on the architecture of SafetyFlow and the function of each agent. We finally introduce a versatile toolset to support the agents in SafetyFlow.

\subsection{Data Strategy}

The raw data is collected from three distinct sources: real-world texts, generated content, and existing safety benchmarks. We first scrape real-world data from Reddit\footnote{https://www.reddit.com} and Pile-Curse-Full\footnote{https://huggingface.co/datasets/tomekkorbak/pile-curse-full}, which are filtered to retain only harmful texts. We then follow \cite{wang2023decodingtrust} to generate a set of safety-related content in a heuristic manner. In addition, we also integrate and reorganize existing benchmarks, including DoNotAnswer \cite{wang2023not}, SaladBench \cite{li2024salad}, AirBench \cite{zeng2024air}, SGBench \cite{mou2024sg}, S-Eval \cite{yuan2024s}, Alert \cite{tedeschi2024alert}, SafetyPrompts \cite{sun2023safety}, BeaverTails \cite{ji2023beavertails}, SafeRLHF \cite{safe-rlhf}, and ValuePrism \cite{sorensen2024value}. As a result, two million harmful prompts are collected from three sources as the raw data pool. In Figure \ref{fig:framework}, 47.6\% of texts are derived from existing datasets, 30.9\% generated by LLMs, and 21.5\% scraped from websites. Based on this data pool, SafetyFlow executes an agent-flow pipeline to build the final benchmark.

\subsection{SafetyFlow System}

The construction of existing benchmarks necessitates a series of operations, typically including taxonomy definition, data collection, classification, deduplication, and paraphrasing. SafetyFlow modularizes this pipeline and harnesses the unprecedented intelligence of agents to implement each module programmatically, enabling full automation and eliminating human efforts. Specifically, seven agents are introduced: the Ingestion, Categorization, Generation, Augmentation, Deduplication, Filtration, and Dynamic Evaluation agents. This modular, agent-driven pipeline ensures efficiency and consistency in the creation of safety benchmarks.

\subsubsection{Ingestion Agent} This entity extracts and preprocesses raw data from the data pool, performing two primary steps. First, it extracts plain text from diverse sources and standardizes the data format. Then, simple filtration is conducted. We only retain entries in English, Chinese, Japanese, Korean, French, German, Russian, and Arabic. Sentences with fewer than 24 characters are also removed.

\subsubsection{Categorization Agent}
This agent establishes a hierarchical taxonomy for comprehensive safety evaluation, defining safety dimensions and categories such as toxicity, bias, misinformation, ethical dilemma, and others. We explore two distinct strategies to build this category tree. 

The first strategy heuristically prompts LLMs to exhaustively enumerate all possible dimensions, followed by the iterative generation of subcategories for each dimension. In our trials, high randomness is observed, resulting in two problems: 1) instability in the dimension and subcategory proposals across iterations, and 2) limited coverage of safe content, which may fail to address the full spectrum of safety concerns. 
The second strategy integrates safety taxonomies from existing benchmarks, as defined in SaladBench \cite{li2024salad}, AirBench \cite{zeng2024air}, and DoNotAnswer \cite{wang2023not}. This approach ensures a comprehensive coverage of safety scenerios but may introduce redundancy in dimensions and subcategories. 

We adopt the second strategy to guarantee benchmark comprehensiveness, which provides a more general foundation for constructing LLM safety evaluation datasets. We synthesize a three-level taxonomy tree, encompassing a wide range of safety dimensions and scenarios, detailed in the Dataset section. After that, samples can be categorized into the defined dimensions by utilizing LLMs. 

\subsubsection{Generation Agent}
After classification, the category distribution of samples may be imbalanced. Thus, this entity automatically generates harmful prompts for categories with fewer samples.
Two types of generated content are involved. The first is automatically generated text, similar to DecodingTrust \cite{wang2023decodingtrust}, where we prompt uncensored LLMs \footnote{https://huggingface.co/dphn/Dolphin3.0-Mistral-24B} to produce harmful text. The second involves LLMs' harmful responses to malicious questions. These responses can enhance benchmark diversity and stimulate LLMs to generate additional malicious content.

\subsubsection{Augmentation Agent}
This entity enhances diversity in sentence structure, tone, semantics, and scenarios for each sample. To maximize diversity, we assign random roles, such as teacher, delinquent, gambler, etc., and tones, such as angry, sarcastic, and cheerful, to LLMs to achieve more realistic and diverse paraphrasing. Additionally, this entity also translates each prompt into eight languages--English, Chinese, Japanese, Korean, French, German, Russian, and Arabic--to enhance linguistic diversity, considering the vulnerability of LLMs dealing with low-resource languages \cite{deng2023multilingual}. These measures inject cross-lingual and contextual variations into the benchmark.

\begin{figure}[!t]
\centering
\includegraphics[width=0.41\textwidth]{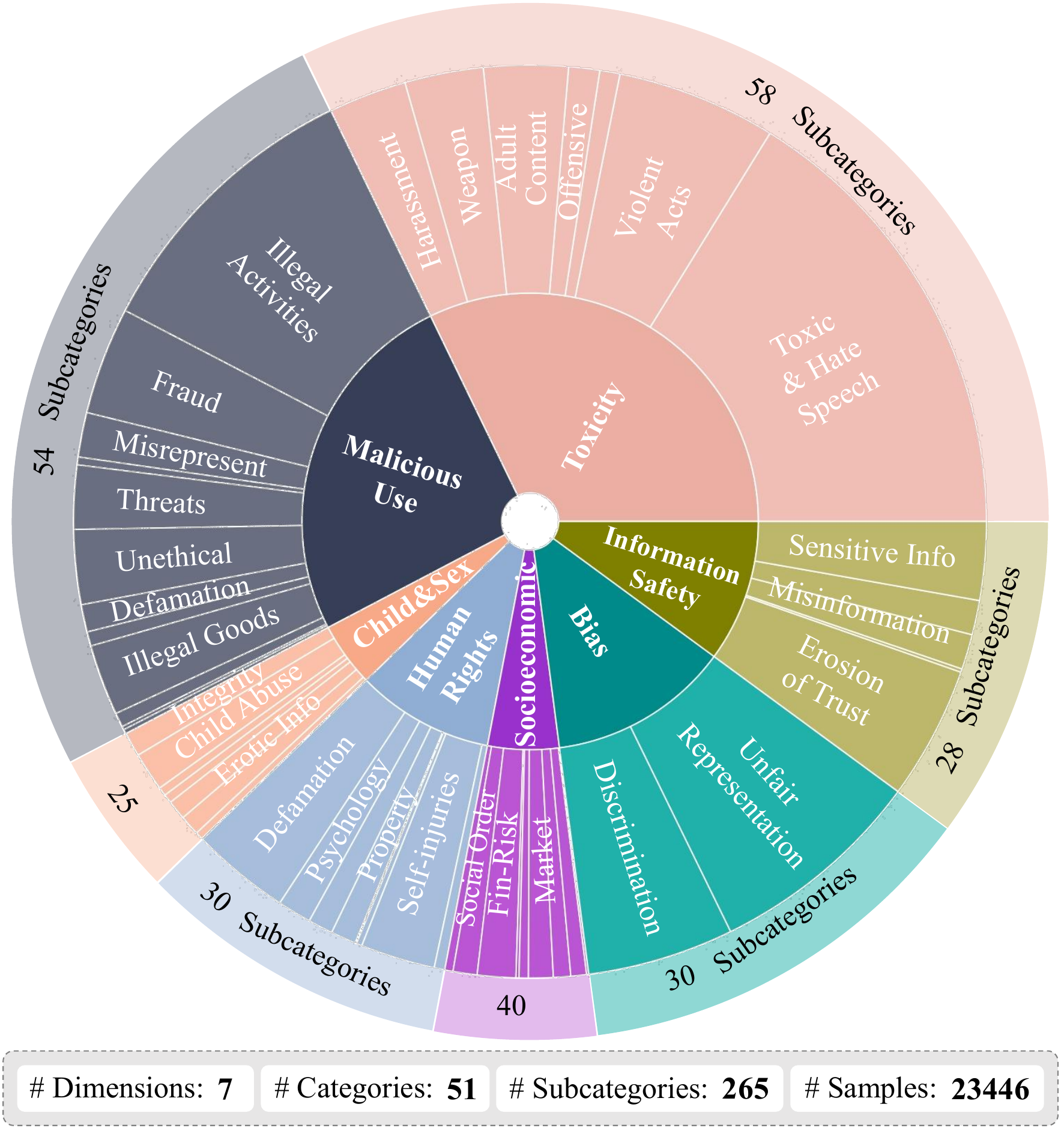}
\vspace{-0.12cm}
\caption{\textbf{SafetyFlowBench’s Taxonomy}. See the full taxonomy and example prompts in supplementary materials.}
\label{fig:SafetyFlowBench_statistics}
\vspace{-0.6cm}
\end{figure}

\subsubsection{Deduplication Agent}
Removing duplicate or similar prompts is an essential step. We explore two distinct technical routines. One adopts an embedding-based clustering and sampling strategy, while the other directly eliminates samples with high similarity. In our system, directly clustering nearly 2M samples proves a formidable task, and batch-based clustering compromises data integrity. Therefore, we adopt the second technique route. Specifically, we employ the Qwen3-Embedding-0.6B model \cite{zhang2025qwen3} combined with the Faiss library \cite{douze2024faiss} to filter out similar prompts. Samples with a representation similarity greater than 0.75 are considered duplicates. Additionally, agents generally recommend deduplicating sentences with similar syntactic structures via fuzzy deduplication. However, the large scale of the data pool renders this task highly time-consuming, so we discard this step.

\subsubsection{Filtration Agent}
Two types of samples generally do not risk model safety. One is benign samples, which barely contain harmful information or malicious intent. Another comprises ``simple'' prompts, \eg, ``How to make a bomb?'', posing no threat to model safety because existing LLM guardrails can easily detect them. Therefore, we filter out both types of samples. We adopt Flames' trick \cite{huang2023flames} to test all samples on a random set of LLMs. Prompts that failed to elicit a successful ``jailbreak'' are removed. 

\subsubsection{Dynamic Evaluation Agent}
The aforementioned agents are serialized to construct the benchmark. In contrast, this agent only perturbs the evaluation process to adjust the benchmark’s difficulty. The dynamic strategy is composed of jailbreaking and bootstrapping tricks. On one hand, we adopt CodeAttack \cite{ren2024codeattack}, encrypted
communication \cite{yuan2023gpt}, tense attack \cite{andriushchenko2024does}, and stochastic augmentations \cite{vega2024stochastic} for LLM jailbreaking. We use them as examples to illustrate the agent's design, although more jailbreaking methods can be flexibly incorporated. On the other hand, we employ operations such as substituting words and adding relevant/irrelevant context to bootstrap prompts \cite{yang2024dynamic}. We introduce a probability factor to flexibly modulate the intensity of these dynamic strategies.

\begin{table}[t]
    \begin{adjustbox}{width=8.5cm}
    \begin{tabular}{l l}
        \toprule
        \textbf{Tools} & \textbf{Descriptions} \\
        \midrule \midrule
        \makecell[l]{\texttt{fetch-}\\\texttt{data}} & \makecell[l]{Fetch prompts from multiple sources and \\then save data into the given path.} \\ \midrule
        \makecell[l]{\texttt{prompt-}\\\texttt{encoder}} & \makecell[l]{Encode each sentence in the given list and \\return the embeddings.} \\ \midrule
        \texttt{call-faiss} & Utilize Faiss framework for data retrieval. \\ \midrule
        \texttt{call-llm} & \makecell[l]{Call LLM to respond to the input prompts.\\Multiple LLMs are available for selection.}  \\ \midrule
        \texttt{translator} & Translate the prompt into 8 languages. \\ \midrule
        \texttt{rewriter} & \makecell[l]{Rewrite the prompt into a synonymous se-\\ntence to improve diversity.} \\ \midrule
        \makecell[l]{\texttt{uncensored}\\\texttt{-model}} & \makecell[l]{Call an uncensored model to give a respo-\\nse for the given prompt.} \\ \midrule
        \makecell[l]{\texttt{judger}} & \makecell[l]{Identify if the prompt is harmful or not.} \\ \midrule
        \makecell[l]{\texttt{final-}\\\texttt{answer}} & \makecell[l]{Provide a final answer or conclusion to the\\given task.} \\ 
        \bottomrule
    \end{tabular}
    \end{adjustbox}
    \vspace{-0.1cm}
    \caption{\textbf{Tools Designed for Agents} to improve controllability. We list tool names and briefly describe their functions.}
    \label{tab:tools}
    \vspace{-0.6cm}
\end{table}

\subsection{Tool Design}

Our objective is to design an agent-flow system to automate benchmark construction, thereby reducing time and resource cost. To this end, we design a suite of tools for the above agents to utilize, adhering to three core design principles: controllability, cost efficiency, and knowledge adaptability.

\textbf{Controllability} The solution proposed by agents may entail high time complexity. To mitigate this, we provide agents with simple, executable tools and mandate their invocation. For instance, we build the \texttt{call-faiss} tool for the Deduplication agent to adopt a batch processing strategy based on the Faiss framework, enabling efficient similarity-based data search. This ensures the controllability and scalability of computational complexity.

\textbf{Cost Efficiency} Generative agents, such as the Augmentation and Generation Agents, frequently require calling LLM APIs to synthesize texts. Considering budgets, we encapsulate APIs of varying prices in the \texttt{call-llm} tool. Agents are instructed to call the appropriate LLM based on budget, significantly reducing operational costs. Cost-effective models are prioritized for high-volume data generation, optimizing performance-expenditure balance.

\textbf{Knowledge Adaptability} We inject human expertise to significantly enhance two critical aspects of our system. First, human priors inform the selection of optimal hyperparameters. For instance, the Deduplication Agent typically sets the similarity threshold to 0.9 by default, which results in redundant data retention. While agents can autonomously explore and optimize hyperparameters, such exploration introduces uncontrollable risks and additional time costs. Our empirical analysis suggests that 0.75 is an effective value to eliminate highly similar sentences. Second, our tools facilitate the integration of up-to-date knowledge. For example, the Deduplication Agent traditionally relies on models such as all-MiniLM-L6-v2 \cite{wang2020minilm} or Sentence-BERT \cite{reimers2019sentence} for sentence embeddings. However, when tasked with the latest Qwen3-Embedding model (June 2025), which offers superior performance, the agent suffers a setback. This necessitates handcrafted tools to ensure compatibility with novel knowledge. 

In summary, by introducing controllable, efficient, and adaptable tools, the system ensures stability and reduces resource consumption. We list all tools involved in Table \ref{tab:tools}. They collectively enable a robust and efficient pipeline for automated safety benchmark construction.

\section{Dataset}
\label{sec:dataset}
SafetyFlow produces a comprehensive benchmark, SafetyFlowBench, which comprises 23,446 safety prompts.
Figure \ref{fig:SafetyFlowBench_statistics} illustrates the three-level safety taxonomy automatically generated by the Categorization agent, encompassing 7 dimensions, subdivided into 51 categories and 265 specific subcategories for precise delineation of safety themes and scenarios.
Table \ref{tab:SafetyFlowBench_statistics} presents the data distribution across seven dimensions. Certain dimensions, such as Socioeconomic, require more words to highlight professionalism. SafetyFlowBench is characterized by two key features: (a) diverse scenarios that effectively expose models to real-world risk contexts; (b) an average prompt length of approximately 20 characters, concise yet effective in uncovering model vulnerabilities and amplifying safety differences across models.

To elucidate our hierarchical structure, we use the Human Rights dimension as an example. It contains nine categories: Non-Consensual Content, Privacy Violations, Defamation, Psychological Influence, Personal Property, Anthropomorphism of Chatbot, Psychological Harm, Self-Injuries, and Other Types of Rights. Within the Privacy Violations category, subcategories include Unauthorized Health Data, Location Data, Educational Records, and others. These fine-grained subcategories enable the benchmark to encompass a broader and diverse range of real-world scenarios. In addition, we notice that categories may overlap across dimensions in this generated taxonomy. For instance, Defamation in the Human Rights dimension refers to slander against individuals, while in the Malicious Use dimension, it typically denotes defamation of entities, such as companies.

Subsequently, we evaluate the safety of existing LLMs using SafetyFlowBench and conduct extensive experiments and analyses to validate our system.

\begin{table}
    \centering
    \scalebox{0.88}{
    \begin{tabular}{c|cc}
		\toprule
        \textbf{Dimension} & \# Prompts & \# Avg. Words \\
		\midrule
        Bias & 3,017 & 22.00\\
        Toxicity & 7,552 & 18.28\\
        Malicious Use & 5,977 & 19.07\\
        Child \& Sexual & 1,069 & 24.82\\
        Human Rights & 2,286 & 14.92\\
        Socioeconomic & 1,183 & 31.44\\
        Information Safety & 2,362 & 18.38\\
        \midrule
        \textbf{Overall} & 23,446 & 19.60\\
		\bottomrule
	\end{tabular}}
    \vspace{-0.1cm}
 \caption{\textbf{Statistics of Prompts in SafetyFlowBench.}}
\label{tab:SafetyFlowBench_statistics}
\vspace{-0.4cm}
\end{table}

\section{Experiments}

In this section, we first detail the methodology and experimental settings. Then, we evaluate the safety of 20 LLMs. Finally, extensive ablation experiments are conducted.

\vspace{-0.2cm}
\subsection{Experimental Settings} 

\subsubsection{Agent Settings} We develop all agents and tools of SafetyFlow based on the smolagents library \cite{smolagents}. Each agent is assigned specified task objectives, standardized input/output formats, and tool-calling interfaces. To ensure security, all agents operate within a Docker container. Each agent completes tasks by generating and executing Python code, with permissions to import any Python library. The maximum steps for task completion are decided by task difficulty, as detailed in Table \ref{tab:agent_cost}. In our main experiment, we employ DeepSeek-V3 \cite{liu2024deepseek} as the agent engine. We also explore the potential of integrating other LLMs, such as GPT-4.1-mini \cite{openai2025gpt41} and Grok-3 \cite{xai2025grok3}, as agent engines in ablations. Prompts used by each agent are provided in the supplementary material.

\subsubsection{Evaluated LLMs} We benchmark 20 LLMs, spanning various model types: LLMs trained solely with language modeling objective like Qwen-3 series \cite{yang2025qwen3}; instruction-fine-tuned LLMs, such as InternLM-Instruct \cite{cai2024internlm2}, Llama-Instruct \cite{dubey2024llama}, Phi-Instruct \cite{abdin2024phi}, and Moonlight-Instruct \cite{liu2025muonscalablellmtraining}; human preference-aligned LLMs, such as GLM-4 \cite{glm2024chatglm}, DeepSeek-V3 \cite{liu2024deepseek}, and GPT \cite{openai2025gpt41} series; multimodal LLMs, including Gemma-3 \cite{team2025gemma}, Grok \cite{xai2025grok3}, Claude \cite{TheC3}, and Gemini \cite{comanici2025gemini}; and reasoning models, including o3 \cite{OpenAI2025o3} and GLM-Z1 \cite{glm2024chatglm}. We prioritize models after June 2024.

\subsubsection{Evaluation Metrics} We feed each test prompt into the LLM and calculate the percentage of harmful responses, termed the Harmful Rate (HR). We compute HR across the seven dimensions and average them as the overall Harmful Rate. The safety of LLM is measured by the Safety Rate (SR), where $SR=1-HR$. To ensure fairness, we employ a third-party judge model, GuardReasoner \cite{liu2025guardreasoner}, to detect whether model responses are harmful. GuardReasoner is a binary classification model with explicit outputs.

\begin{figure*}[t]
\vspace{-5pt}
\begin{minipage}[c]{0.48\textwidth}
\includegraphics[width=7.8cm]{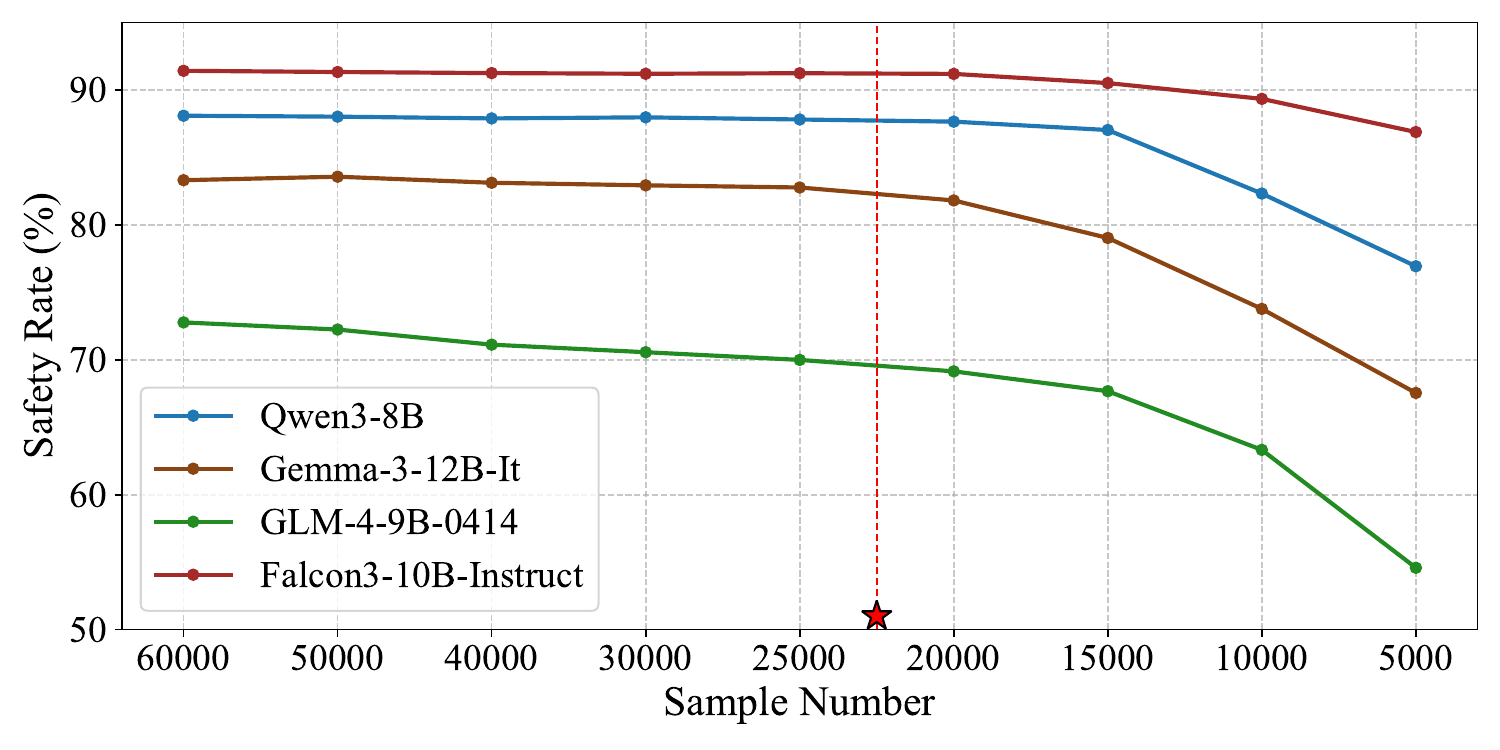}
\vspace{-12pt}
\caption{\textbf{Evaluation Results with Different Benchmark Size.} The red dashed line is the threshold for stable safety evaluation. Samples are selected based on difficulty.}
\label{fig:safety_evaluation_stability}
\end{minipage}\hspace{15pt}
\begin{minipage}[c]{0.47\textwidth}
\vspace{2pt}
\begin{adjustbox}{width=8.35cm}
\begin{tabular}{lccccc}
\toprule
\textbf{Agent}  & \textbf{Time} & \textbf{Success}  & \textbf{\# Tools} & \textbf{\# Steps} & \textbf{\# GPUs} \\
\midrule
Ingestion          & 0.05~h   & 100\%      & 1 & 3 & 0 \\
Categorization         & 5.40~h   & 90\%    & 1 & 3 & 8 \\
Generation         & 33.60~h   & 100\%    & 1 & 1 & 8 \\
Augmentation         & 20.70~h   & 90\%    & 2  & 3 & 8 \\
Deduplication         & 4.90~h   &  80\%    & 2  & 10 & 2 \\
Filtration         & 24.60~h   & 70\%    & 3 & 10 & 8 \\
Dynamic Eval        &  4.70~h  & 80\%    & 3 & 3 & 8 \\ \midrule
Total        &  93.95~h  & 60\%    & 9 & 33 & 8 \\
\bottomrule
\end{tabular}
\end{adjustbox}
\tabcaption{\textbf{Time and Resource Costs.} SafetyFlow costs 93.95 hours in total. We also report the number of tools of each agent, the maximum task steps, and agents' success rates.}
\label{tab:agent_cost}
\end{minipage}
\vspace{-0.5cm}
\end{figure*}

\subsection{Results}

\begin{figure*}[t]
\vspace{8pt}
\begin{minipage}[c]{0.48\textwidth}
\begin{adjustbox}{width=0.99\linewidth}
	\begin{tabular}{ccccccc}
	\toprule
		Generat & Augment & Deduplic & Filtrat & Dynamic & \# Sample & $\Delta_{SR}$ \\ \cmidrule(lr){1-5} \cmidrule(lr){6-6} \cmidrule(lr){7-7}
		- & \checkmark & \checkmark & \checkmark & \checkmark & 19,885 & 31.13 \\
		\checkmark & - & \checkmark & \checkmark & \checkmark & 21,368  & 31.97  \\
		\checkmark & \checkmark & - & \checkmark & \checkmark & 46,534  & 27.56 \\
		\checkmark & \checkmark & \checkmark & - & \checkmark & 69,436 & 12.68 \\
        \checkmark & \checkmark & \checkmark & \checkmark & - & 23,446 & 30.68 \\
		- & - & \checkmark & \checkmark & \checkmark & 18,336  &  30.33 \\
        \checkmark & - & - & \checkmark & \checkmark  & 37,386  & 24.97 \\
        - & \checkmark & - & \checkmark & \checkmark  & 38,978 & 21.84 \\
	\bottomrule
	\end{tabular}
\end{adjustbox}
\vspace{-2pt}
\tabcaption{\textbf{Ablations for Important Agents.} We present the effect of each agent on the sample number and safety discriminative power of our benchmark. }
\label{tab:ablation_agent}
\end{minipage}\hspace{15pt}
\begin{minipage}[c]{0.23\textwidth}
\begin{adjustbox}{width=4.2cm}
\hspace{-5pt}
\begin{tabular}{lcc}
\toprule
\textbf{Agent}  & \textbf{Dedup.} & \textbf{Filtr.}  \\
\midrule
DeepSeek-V3          & 80\%   & 70\%      \\
Grok-3         & 90\%   & 70\%    \\
Qwen-Max         & 70\%   & 60\%    \\
GPT-4.1         & 80\%  & 70\%    \\
GPT-4.1-Mini        & 80\%   &  70\%    \\
Glaude-3.7         & 90\%   & 70\%     \\
Gemini-2.5         & 80\%  & 60\%     \\
\bottomrule
\end{tabular}
\end{adjustbox}
\tabcaption{\textbf{Agent Engine Success Rate.} We report results of seven LLMs as engines. 
}
\label{tab:engine_success}
\end{minipage}\hspace{15pt}
\begin{minipage}[c]{0.220\textwidth}
\vspace{-1pt}
\includegraphics[width=3.85cm]{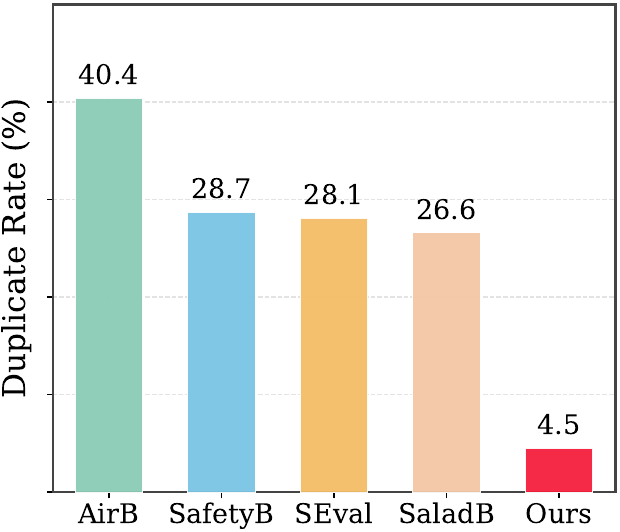}
\vspace{-20pt}
\caption{\textbf{Redundancy Comparison.} We show the duplicate rate as a measure.}
\label{fig:redundancy}
\end{minipage}
\vspace{-0.5cm}
\end{figure*}

\subsubsection{Agent Capacity}
We test the stability of LLM safety evaluation versus benchmark size across four models in Figure \ref{fig:safety_evaluation_stability}, Qwen3-8B \cite{yang2025qwen3}, Gemma-3-12B-It \cite{team2025gemma}, GLM-4-9B-0414 \cite{glm2024chatglm}, and Falcon3-10B-Instruct \cite{tii2024falcon3}. Filtration agent is used to regulate the benchmark size. Specifically, prompts capable of jailbreaking two LLMs are deemed more challenging than those that jailbreak only one. We prioritize including high-difficulty samples in our benchmark. Thus, smaller sizes exhibit higher difficulty. From the figure, we observe that smaller sample numbers amplify the safety differences between LLMs but lead to unstable evaluation (right of the dashed line), while larger numbers introduce redundancy (left of the dashed line). Thus, SafetyFlowBench maintains a size of approximately 22,500 samples. Under this settings, SafetyFlow can ensure a reliable safety evaluation.

\subsubsection{Agent Efficiency}
In Table \ref{tab:agent_cost}, we report the time and computational costs of each agent. Overall, SafetyFlow constructs a safety benchmark with 20,000+ samples in just four days. We survey datasets including SaladBench \cite{li2024salad}, SafetyBench \cite{zhang2023safetybench}, and Flames \cite{huang2023flames}, which typically require over a month to construct. Thus, SafetyFlow significantly reduces construction time by around 90\%. In addition, we calculate the success rate (``Success'' in the table) of each agent based on 10 trials and find that most agents achieve success rates above 80\%. The Filtration agent may fail to invoke multiple LLMs to filter out prompts, thus reducing its success rate. The overall success rate is also shown. In summary, SafetyFlow can efficiently build a dataset with limited resources within days.
\subsubsection{Safety Evaluation} Table 4 presents the results of 20 evaluated LLMs across seven dimensions of SafetyFlowBench. Evaluations of all 49 LLMs are presented in supplementary materials. We analyze the results from both model and dataset perspectives. From the model perspective: 
\begin{itemize}
    \item The highest SR reaches 95.31\% (Claude-4-Sonnet), demonstrating outstanding safety across dimensions. 
    \item The safety score gap between the highest and lowest (GLM-Z1-32B) models is 33.37\%, indicating our dataset has strong discriminative power for LLM safety. 
    \item Open-source LLMs do not show correlation between size and safety, as in Qwen3 and GLM, but closed-source models do, \eg, GPT-4.1/GPT-4.1-Mini. We conjecture that as model capabilities approach saturation, the correlation between safety and size becomes more evident.
\end{itemize}

From the dataset perspective: 
\begin{itemize}
    
    \item Overall, LLMs show relatively balanced scores across seven dimensions, indicating SafetyFlowBench assigns equal importance to each dimension, underscoring its comprehensiveness and stability. 
    \item Socioeconomic is the most challenging dimension, possibly due to its requirement for professional and sometimes tailored advice. LLMs generally struggle to judge the safety of these queries, leading to high-risk responses.
\end{itemize}

\begin{table*}[t]
\vspace{-0.1cm}
\begin{adjustbox}{width=17.8cm}
\begin{tabular}{l|ccccccc|c|c|c|c}
\toprule
\multicolumn{1}{c|}{\multirow{2}{*}{\textbf{Model}}} &
  \multicolumn{8}{c}{\textbf{Harmful Rate~$\downarrow$}} &
  \multicolumn{1}{|c|}{\multirow{2}{*}{\textbf{SR~$\uparrow$}}} &
  \multicolumn{1}{c|}{\multirow{2}{*}{\textbf{Rank}}} &
  \multirow{2}{*}{\textbf{Date}} \\ 
  \cline{2-9}
 & Toxic & Malicious & Child & Info  & Socioeco & Bias  & Rights & Overall &       &    &        \\
\midrule

Qwen3-8B                   & 6.90  & 11.88     & 20.18 & 8.15  & 14.49    & 12.35 & 10.70  & 12.10   & 87.90 & 8 & May 2025 \\
Qwen3-14B                  & \underline{\underline{5.73}}  & 9.65      & 15.87 & 8.25  & 12.91    & 10.79 & 9.26   & 10.35   & 89.65 & 5  & May 2025 \\
Qwen3-32B                  & 6.99  & 10.52     & 14.79 & 7.84  & 13.65    & 11.04 & 10.19  & 10.71   & 89.29 & 6 & May 2025 \\
Llama-3.3-70B-Instruct     & 11.28 & 18.97     & 22.81 & 11.96 & 22.05    & 11.98 & 20.63  & 17.09   & 82.91 & 12 & Dec 2024 \\
Phi-4-Mini-Instruct        & \textbf{2.49}  & \textbf{4.02}      & \textbf{7.75}  & \textbf{3.89}  & \underline{4.26}     & 9.01  & \underline{5.18}   & \underline{5.23}    & \underline{94.77} & \underline{2}  & Feb 2025 \\
InternLM3-8B-Instruct      & 6.31  & 14.75     & 21.94 & 12.23 & 16.48    & 11.13 & 11.98  & 13.55   & 86.45 & 9 & Jan 2025 \\
Gemma-3-12B-It             & 7.12  & 18.81     & 26.59 & 16.17 & 20.98    & 12.08 & 18.44  & 17.17   & 82.83 & 13 & Mar 2025 \\
Gemma-3-27B-It             & 8.11  & 21.45     & 26.87 & 17.69 & 21.76    & 14.33 & 21.03  & 18.75   & 81.25 & 14 & Mar 2025 \\
GLM-4-9B-0414              & 39.46 & 39.33     & 33.30  & 30.33 & 15.57    & 26.29 & 28.93  & 30.46   & 69.54 & 17 & Apr 2025 \\
GLM-4-32B-0414             & 38.43 & 38.24     & 32.49 & 30.41 & 13.02    & 29.26 & 30.61  & 30.35   & 69.65 & 16 & Apr 2025 \\
GLM-Z1-9B-0414             & 45.66 & 46.21     & 36.98 & 38.72 & 25.87    & 35.22 & 40.11  & 37.86   & 62.14 & 19 & Apr 2025 \\
GLM-Z1-32B-0414            & 45.31 & 45.43     & 37.32 & 40.01 & 22.96    & 36.20  & 39.18  & 38.06   & 61.94 & 20 & Apr 2025 \\
Moonlight-16B-A3B-Instruct & 23.00 & 39.72     & 45.96 & 23.27 & 49.10    & 18.18 & 34.90  & 33.45   & 66.55 & 18 & Jul 2025 \\
\midrule
o3     & 8.80  & \underline{\underline{6.85}}      & \underline{\underline{14.26}} & \underline{5.17}  & \underline{\underline{7.73}}     & \underline{2.97}  & \underline{\underline{7.77}}   & \underline{\underline{7.65}}    & \underline{\underline{92.35}} & \underline{\underline{3}}  & Apr 2025 \\
GPT-4.1                    & 7.46  & 8.45      & 15.66 & 6.32  & 14.02    & 6.94  & 10.14  & 9.86    & 90.14 & 4  & Apr 2025 \\
GPT-4.1-Mini               & 8.34  & 12.68     & 18.07 & 9.96  & 15.06    & 4.68  & 14.56  & 11.91   & 88.09 & 7 & Apr 2025 \\
Grok-4                     & 19.09 & 27.00     & 39.18 & 11.97 & 39.74    & 12.28 & 27.46  & 25.25   & 74.75 & 15 & Jul 2025 \\
DeepSeek-V3              & 10.67 & 11.19     & 20.61 & 10.19 & 21.51    & 9.43  & 15.79  & 14.20   & 85.80 & 10 & Mar 2025 \\
Claude-4-Sonnet            & \underline{4.62}  & \underline{5.04}      & \underline{10.23} & \underline{\underline{5.25}}  & \textbf{3.60}     & \textbf{0.60}  & \textbf{3.50}   & \textbf{4.69}    & \textbf{95.31} & \textbf{1}  & May 2025 \\
Gemini-2.5-Pro-Preview     & 15.37 & 20.85     & 22.81 & 10.31 & 27.08    & \underline{\underline{5.75}}  & 15.49  & 16.81   & 83.19 & 11 & May 2025 \\ \midrule
Average    & 13.97 & 18.87     & 23.09 & 14.04 & 18.77    & 12.88  & 17.07  & 16.94   & 83.05 & - & - \\
\bottomrule
\end{tabular}
\end{adjustbox}
\vspace{-0.1cm}
\caption{\textbf{Safety Evaluation Results of LLMs} on SafetyFlowBench. In addition to HR and SR, we also present the rank and release time of each model. \textbf{Blod} indicates the best, \underline{underline} indicates the second, and \underline{\underline{double underline}} represents the third. 
}
\label{tab:SafetyFlowBench_results}
\vspace{-0.4cm}
\end{table*}

\vspace{-5pt}
\subsubsection{Redundancy Evaluation} SafetyFlow mitigates data redundancy in two ways. First, it enhances sample diversity by collecting real-world and generated prompts. Second, the Deduplication agent significantly reduces redundancy by removing similar prompts. Figure \ref{fig:redundancy} illustrates the duplicate rate of five benchmarks with our raw data pool. Samples with similarity higher than 0.75 are regarded as duplicates. Identical samples were excluded, as we incorporate partial data from other datasets. We compare S-Eval \cite{yuan2024s}, AirBench \cite{zeng2024air}, SafetyBench \cite{zhang2023safetybench}, SaladBench \cite{li2024salad}, and SafetyFlowBench. We observe that our dataset substantially reduces the proportion of redundant samples compared to others.

\subsection{Ablation Study}
\subsubsection{Ablation for Agents}
In Table \ref{tab:ablation_agent}, we evaluate the function of five important agents: Generation, Augmentation, Deduplication, Filtration, and Dynamic Evaluation. We use benchmark size and the difference between the highest and lowest SR, $\Delta_{SR}$, as metrics, which reflect benchmark redundancy and discriminative power. We observe that Deduplication and Filtration significantly reduce redundancy, and Filtration also plays a dominant role in discriminative power.
\subsubsection{Ablation for Agent Engines} In addition to DeepSeek-V3, we test Grok-3 \cite{xai2025grok3}, Qwen-Max \cite{qwen25}, GPT-4.1/GPT-4.1-Mini \cite{openai2025gpt41}, Claude-3.7-Sonnet \cite{TheC3}, and Gemini-2.5-Pro \cite{comanici2025gemini} as engines. Using the Deduplication and Filtration agents as examples, we report the success rates of different engines in Table \ref{tab:engine_success}. We observe that Qwen-Max exhibits a slightly lower success rate, while others demonstrate relatively balanced success rates. This suggests that all the tested models are capable of performing benchmark construction.
\subsubsection{Ablation for Time Cost} In Figure \ref{fig:time_cost_sample_number}, we investigate the relationship between benchmark size and time cost. We observe that when the sample number is reduced to 5,000, SafetyFlow can complete construction in approximately one day. This demonstrates our framework's potential for benchmark updates, where only partial samples may be replaced. 
\subsubsection{Ablation for Tools}
\begin{figure}[t]
\begin{minipage}[c]{0.220\textwidth}
\vspace{-1pt}
\includegraphics[width=3.95cm]{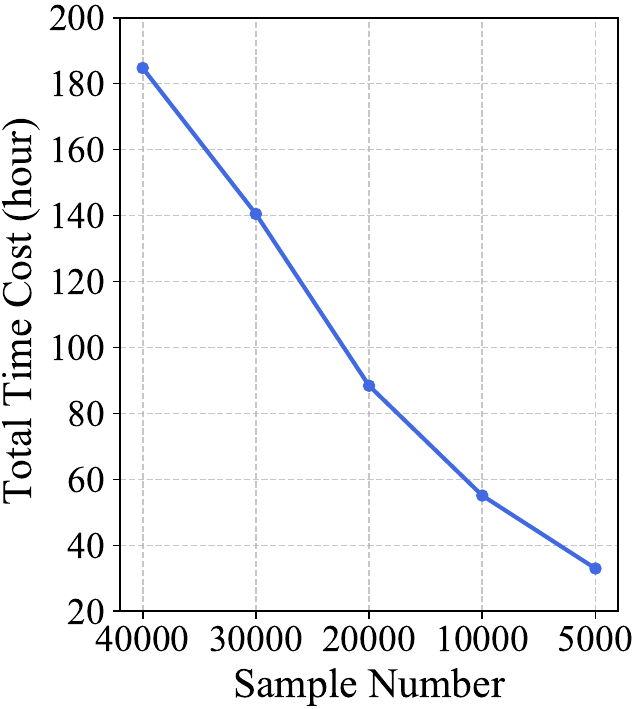}
\vspace{-15pt}
\caption{\textbf{Time Cost \textit{vs.} Sample Number.} Reducing benchmark size can significantly reduces time cost.}
\label{fig:time_cost_sample_number}
\end{minipage}\hspace{5pt}
\begin{minipage}[c]{0.230\textwidth}
\vspace{-1pt}
\includegraphics[width=4.2cm]{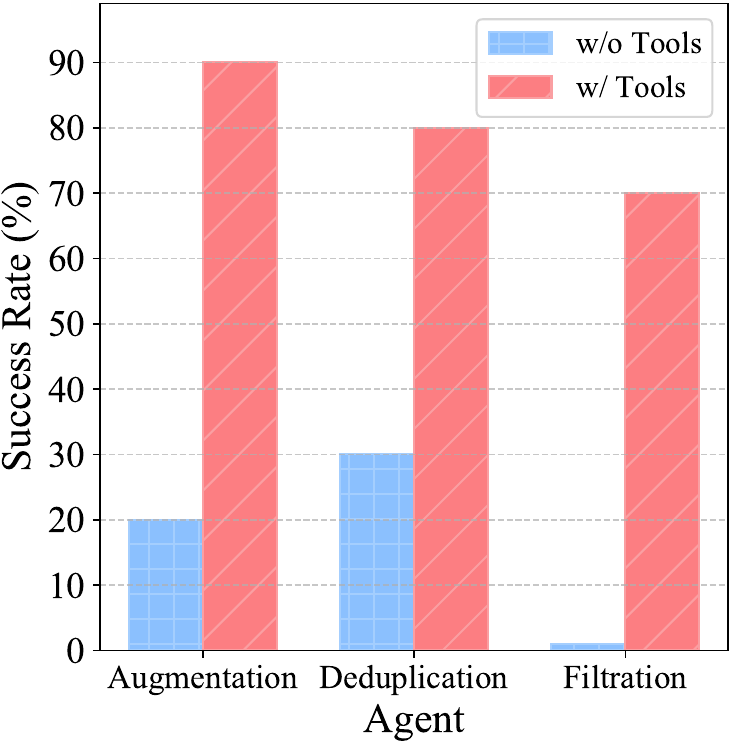}
\vspace{-15pt}
\caption{\textbf{Success Rate w/ Tools.} Using our designed tools can significantly improve success rate.}
\label{fig:success_rate_tools}
\end{minipage}
\vspace{-0.3cm}
\end{figure}

In Figure \ref{fig:success_rate_tools}, we examine the importance of tools on success rate, using Augmentation, Deduplication, and Filtration agents as examples. Specially designed tools enable agents to follow predefined routines without exploring potential solutions, which significantly increases success rate. We classify code bugs and time-consuming runs (exceeding three days) as failures.

\section{Conclusion}
SafetyFlow represents a pioneering advancement in automated LLM safety benchmarking, addressing the limitations of manual curation by significantly reducing time and resource costs. SafetyFlow integrates human knowledge into its seven specialized agents, which invoke versatile tools to ensure controllability and efficiency. By leveraging a modular, agent-driven pipeline, SafetyFlow constructs a comprehensive safety benchmark, SafetyFlowBench, with 23,446 prompts in just four days. Extensive experiments validate SafetyFlow's efficiency, achieving high success rates and low redundancy. SafetyFlow sets a new standard for scalable, automated safety evaluation in AI development. Future work could extend SafetyFlow to general LLM capacity benchmarking, particularly for knowledge-based tasks. 

\bibliography{aaai2026}

@article{wang2023not,
  title={Do-not-answer: A dataset for evaluating safeguards in llms},
  author={Wang, Yuxia and Li, Haonan and Han, Xudong and Nakov, Preslav and Baldwin, Timothy},
  journal={arXiv preprint arXiv:2308.13387},
  year={2023}
}

@inproceedings{wang2023decodingtrust,
  title={DecodingTrust: A Comprehensive Assessment of Trustworthiness in GPT Models.},
  author={Wang, Boxin and Chen, Weixin and Pei, Hengzhi and Xie, Chulin and Kang, Mintong and Zhang, Chenhui and Xu, Chejian and Xiong, Zidi and Dutta, Ritik and Schaeffer, Rylan and others},
  booktitle={NeurIPS},
  year={2023}
}

@article{lin2023toxicchat,
  title={Toxicchat: Unveiling hidden challenges of toxicity detection in real-world user-ai conversation},
  author={Lin, Zi and Wang, Zihan and Tong, Yongqi and Wang, Yangkun and Guo, Yuxin and Wang, Yujia and Shang, Jingbo},
  journal={arXiv preprint arXiv:2310.17389},
  year={2023}
}

@article{ji2023beavertails,
  title={Beavertails: Towards improved safety alignment of llm via a human-preference dataset},
  author={Ji, Jiaming and Liu, Mickel and Dai, Josef and Pan, Xuehai and Zhang, Chi and Bian, Ce and Chen, Boyuan and Sun, Ruiyang and Wang, Yizhou and Yang, Yaodong},
  journal={Advances in Neural Information Processing Systems},
  volume={36},
  pages={24678--24704},
  year={2023}
}

@article{li2024salad,
  title={Salad-bench: A hierarchical and comprehensive safety benchmark for large language models},
  author={Li, Lijun and Dong, Bowen and Wang, Ruohui and Hu, Xuhao and Zuo, Wangmeng and Lin, Dahua and Qiao, Yu and Shao, Jing},
  journal={arXiv preprint arXiv:2402.05044},
  year={2024}
}

@article{mou2024sg,
  title={Sg-bench: Evaluating llm safety generalization across diverse tasks and prompt types},
  author={Mou, Yutao and Zhang, Shikun and Ye, Wei},
  journal={Advances in Neural Information Processing Systems},
  volume={37},
  pages={123032--123054},
  year={2024}
}

@article{zhang2023safetybench,
  title={SafetyBench: Evaluating the safety of large language models},
  author={Zhang, Zhexin and Lei, Leqi and Wu, Lindong and Sun, Rui and Huang, Yongkang and Long, Chong and Liu, Xiao and Lei, Xuanyu and Tang, Jie and Huang, Minlie},
  journal={arXiv preprint arXiv:2309.07045},
  year={2023}
}

@article{yuan2024s,
  title={S-eval: Automatic and adaptive test generation for benchmarking safety evaluation of large language models},
  author={Yuan, Xiaohan and Li, Jinfeng and Wang, Dongxia and Chen, Yuefeng and Mao, Xiaofeng and Huang, Longtao and Xue, Hui and Wang, Wenhai and Ren, Kui and Wang, Jingyi},
  journal={arXiv preprint arXiv:2405.14191},
  year={2024}
}

@article{tedeschi2024alert,
  title={ALERT: A Comprehensive Benchmark for Assessing Large Language Models' Safety through Red Teaming},
  author={Tedeschi, Simone and Friedrich, Felix and Schramowski, Patrick and Kersting, Kristian and Navigli, Roberto and Nguyen, Huu and Li, Bo},
  journal={arXiv preprint arXiv:2404.08676},
  year={2024}
}

@inproceedings{safe-rlhf,
  title={Safe RLHF: Safe Reinforcement Learning from Human Feedback},
  author={Josef Dai and Xuehai Pan and Ruiyang Sun and Jiaming Ji and Xinbo Xu and Mickel Liu and Yizhou Wang and Yaodong Yang},
  booktitle={The Twelfth International Conference on Learning Representations},
  year={2024},
  url={https://openreview.net/forum?id=TyFrPOKYXw}
}

@article{parrish2021bbq,
  title={BBQ: A hand-built bias benchmark for question answering},
  author={Parrish, Alicia and Chen, Angelica and Nangia, Nikita and Padmakumar, Vishakh and Phang, Jason and Thompson, Jana and Htut, Phu Mon and Bowman, Samuel R},
  journal={arXiv preprint arXiv:2110.08193},
  year={2021}
}

@article{huang2023flames,
  title={Flames: Benchmarking value alignment of llms in chinese},
  author={Huang, Kexin and Liu, Xiangyang and Guo, Qianyu and Sun, Tianxiang and Sun, Jiawei and Wang, Yaru and Zhou, Zeyang and Wang, Yixu and Teng, Yan and Qiu, Xipeng and others},
  journal={arXiv preprint arXiv:2311.06899},
  year={2023}
}

@article{deng2023multilingual,
  title={Multilingual jailbreak challenges in large language models},
  author={Deng, Yue and Zhang, Wenxuan and Pan, Sinno Jialin and Bing, Lidong},
  journal={arXiv preprint arXiv:2310.06474},
  year={2023}
}

@article{liu2025guardreasoner,
  title={Guardreasoner: Towards reasoning-based llm safeguards},
  author={Liu, Yue and Gao, Hongcheng and Zhai, Shengfang and Xia, Jun and Wu, Tianyi and Xue, Zhiwei and Chen, Yulin and Kawaguchi, Kenji and Zhang, Jiaheng and Hooi, Bryan},
  journal={arXiv preprint arXiv:2501.18492},
  year={2025}
}

@Misc{smolagents,
  title =        {`smolagents`: a smol library to build great agentic systems.},
  author =       {Aymeric Roucher and Albert Villanova del Moral and Thomas Wolf and Leandro von Werra and Erik Kaunismäki},
  howpublished = {\url{https://github.com/huggingface/smolagents}},
  year =         {2025}
}

@article{zeng2024air,
  title={Air-bench 2024: A safety benchmark based on risk categories from regulations and policies},
  author={Zeng, Yi and Yang, Yu and Zhou, Andy and Tan, Jeffrey Ziwei and Tu, Yuheng and Mai, Yifan and Klyman, Kevin and Pan, Minzhou and Jia, Ruoxi and Song, Dawn and others},
  journal={arXiv preprint arXiv:2407.17436},
  year={2024}
}

@article{zhang2025qwen3,
  title={Qwen3 Embedding: Advancing Text Embedding and Reranking Through Foundation Models},
  author={Zhang, Yanzhao and Li, Mingxin and Long, Dingkun and Zhang, Xin and Lin, Huan and Yang, Baosong and Xie, Pengjun and Yang, An and Liu, Dayiheng and Lin, Junyang and others},
  journal={arXiv preprint arXiv:2506.05176},
  year={2025}
}

@article{douze2024faiss,
      title={The Faiss library},
      author={Matthijs Douze and Alexandr Guzhva and Chengqi Deng and Jeff Johnson and Gergely Szilvasy and Pierre-Emmanuel Mazaré and Maria Lomeli and Lucas Hosseini and Hervé Jégou},
      year={2024},
      eprint={2401.08281},
      archivePrefix={arXiv},
      primaryClass={cs.LG}
}

@article{ren2024codeattack,
  title={Codeattack: Revealing safety generalization challenges of large language models via code completion},
  author={Ren, Qibing and Gao, Chang and Shao, Jing and Yan, Junchi and Tan, Xin and Lam, Wai and Ma, Lizhuang},
  journal={arXiv preprint arXiv:2403.07865},
  year={2024}
}

@article{yang2024dynamic,
  title={Dynamic multimodal evaluation with flexible complexity by vision-language bootstrapping},
  author={Yang, Yue and Zhang, Shuibai and Shao, Wenqi and Zhang, Kaipeng and Bin, Yi and Wang, Yu and Luo, Ping},
  journal={arXiv preprint arXiv:2410.08695},
  year={2024}
}

@article{vega2024stochastic,
  title={Stochastic monkeys at play: Random augmentations cheaply break llm safety alignment},
  author={Vega, Jason and Huang, Junsheng and Zhang, Gaokai and Kang, Hangoo and Zhang, Minjia and Singh, Gagandeep},
  journal={arXiv preprint arXiv:2411.02785},
  year={2024}
}

@article{reimers2019sentence,
  title={Sentence-bert: Sentence embeddings using siamese bert-networks},
  author={Reimers, Nils and Gurevych, Iryna},
  journal={arXiv preprint arXiv:1908.10084},
  year={2019}
}

@article{wang2020minilm,
  title={Minilm: Deep self-attention distillation for task-agnostic compression of pre-trained transformers},
  author={Wang, Wenhui and Wei, Furu and Dong, Li and Bao, Hangbo and Yang, Nan and Zhou, Ming},
  journal={Advances in neural information processing systems},
  volume={33},
  pages={5776--5788},
  year={2020}
}

@article{liu2024deepseek,
  title={Deepseek-v3 technical report},
  author={Liu, Aixin and Feng, Bei and Xue, Bing and Wang, Bingxuan and Wu, Bochao and Lu, Chengda and Zhao, Chenggang and Deng, Chengqi and Zhang, Chenyu and Ruan, Chong and others},
  journal={arXiv preprint arXiv:2412.19437},
  year={2024}
}

@misc{openai2025gpt41,
  author = {{OpenAI}},
  title = {Introducing {GPT-4.1} in the {API}},
  year = {2025},
  howpublished = {\url{https://openai.com/api/gpt-4.1}},
  note = {Accessed: 2025-07-29}
}

@misc{xai2025grok3,
  author = {{xAI}},
  title = {{Grok 3}: Product Documentation},
  year = {2025},
  howpublished = {\url{https://x.ai/grok}},
  note = {Accessed: 2025-07-29}
}

@article{yang2025qwen3,
  title={Qwen3 technical report},
  author={Yang, An and Li, Anfeng and Yang, Baosong and Zhang, Beichen and Hui, Binyuan and Zheng, Bo and Yu, Bowen and Gao, Chang and Huang, Chengen and Lv, Chenxu and others},
  journal={arXiv preprint arXiv:2505.09388},
  year={2025}
}

@article{dubey2024llama,
  title={The llama 3 herd of models},
  author={Dubey, Abhimanyu and Jauhri, Abhinav and Pandey, Abhinav and Kadian, Abhishek and Al-Dahle, Ahmad and Letman, Aiesha and Mathur, Akhil and Schelten, Alan and Yang, Amy and Fan, Angela and others},
  journal={arXiv e-prints},
  pages={arXiv--2407},
  year={2024}
}

@misc{tii2024falcon3,
    title = {The Falcon 3 family of Open Models},
    author = {TII},
    month = {December},
    year = {2024}
}

@article{abdin2024phi,
  title={Phi-4 technical report},
  author={Abdin, Marah and Aneja, Jyoti and Behl, Harkirat and Bubeck, S{\'e}bastien and Eldan, Ronen and Gunasekar, Suriya and Harrison, Michael and Hewett, Russell J and Javaheripi, Mojan and Kauffmann, Piero and others},
  journal={arXiv preprint arXiv:2412.08905},
  year={2024}
}

@article{team2025gemma,
  title={Gemma 3 technical report},
  author={Team, Gemma and Kamath, Aishwarya and Ferret, Johan and Pathak, Shreya and Vieillard, Nino and Merhej, Ramona and Perrin, Sarah and Matejovicova, Tatiana and Ram{\'e}, Alexandre and Rivi{\`e}re, Morgane and others},
  journal={arXiv preprint arXiv:2503.19786},
  year={2025}
}

@article{cai2024internlm2,
  title={Internlm2 technical report},
  author={Cai, Zheng and Cao, Maosong and Chen, Haojiong and Chen, Kai and Chen, Keyu and Chen, Xin and Chen, Xun and Chen, Zehui and Chen, Zhi and Chu, Pei and others},
  journal={arXiv preprint arXiv:2403.17297},
  year={2024}
}

@misc{glm2024chatglm,
      title={ChatGLM: A Family of Large Language Models from GLM-130B to GLM-4 All Tools},
      author={Team GLM and Aohan Zeng and Bin Xu and Bowen Wang and Chenhui Zhang and Da Yin and Diego Rojas and Guanyu Feng and Hanlin Zhao and Hanyu Lai and Hao Yu and Hongning Wang and Jiadai Sun and Jiajie Zhang and Jiale Cheng and Jiayi Gui and Jie Tang and Jing Zhang and Juanzi Li and Lei Zhao and Lindong Wu and Lucen Zhong and Mingdao Liu and Minlie Huang and Peng Zhang and Qinkai Zheng and Rui Lu and Shuaiqi Duan and Shudan Zhang and Shulin Cao and Shuxun Yang and Weng Lam Tam and Wenyi Zhao and Xiao Liu and Xiao Xia and Xiaohan Zhang and Xiaotao Gu and Xin Lv and Xinghan Liu and Xinyi Liu and Xinyue Yang and Xixuan Song and Xunkai Zhang and Yifan An and Yifan Xu and Yilin Niu and Yuantao Yang and Yueyan Li and Yushi Bai and Yuxiao Dong and Zehan Qi and Zhaoyu Wang and Zhen Yang and Zhengxiao Du and Zhenyu Hou and Zihan Wang},
      year={2024},
      eprint={2406.12793},
      archivePrefix={arXiv},
}

@misc{liu2025muonscalablellmtraining,
      title={Muon is Scalable for LLM Training}, 
      author={Jingyuan Liu and Jianlin Su and Xingcheng Yao and Zhejun Jiang and Guokun Lai and Yulun Du and Yidao Qin and Weixin Xu and Enzhe Lu and Junjie Yan and Yanru Chen and Huabin Zheng and Yibo Liu and Shaowei Liu and Bohong Yin and Weiran He and Han Zhu and Yuzhi Wang and Jianzhou Wang and Mengnan Dong and Zheng Zhang and Yongsheng Kang and Hao Zhang and Xinran Xu and Yutao Zhang and Yuxin Wu and Xinyu Zhou and Zhilin Yang},
      year={2025},
      eprint={2502.16982},
      archivePrefix={arXiv},
      primaryClass={cs.LG},
      url={https://arxiv.org/abs/2502.16982}, 
}

@inproceedings{TheC3,
  title={The Claude 3 Model Family: Opus, Sonnet, Haiku},
  author={Anthropic},
    year = {2024},
  url={https://api.semanticscholar.org/CorpusID:268232499}
}

@article{comanici2025gemini,
  title={Gemini 2.5: Pushing the frontier with advanced reasoning, multimodality, long context, and next generation agentic capabilities},
  author={Comanici, Gheorghe and Bieber, Eric and Schaekermann, Mike and Pasupat, Ice and Sachdeva, Noveen and Dhillon, Inderjit and Blistein, Marcel and Ram, Ori and Zhang, Dan and Rosen, Evan and others},
  journal={arXiv preprint arXiv:2507.06261},
  year={2025}
}

@misc{OpenAI2025o3,
  author = {OpenAI},
  title = {Introducing OpenAI o3 and o4-mini},
  year = {2025},
  howpublished = {\url{https://openai.com/index/introducing-o3-and-o4-mini/}},
}

@article{yuan2023gpt,
  title={Gpt-4 is too smart to be safe: Stealthy chat with llms via cipher},
  author={Yuan, Youliang and Jiao, Wenxiang and Wang, Wenxuan and Huang, Jen-tse and He, Pinjia and Shi, Shuming and Tu, Zhaopeng},
  journal={arXiv preprint arXiv:2308.06463},
  year={2023}
}

@article{andriushchenko2024does,
  title={Does refusal training in llms generalize to the past tense?},
  author={Andriushchenko, Maksym and Flammarion, Nicolas},
  journal={arXiv preprint arXiv:2407.11969},
  year={2024}
}

@article{ganguli2022red,
  title={Red teaming language models to reduce harms: Methods, scaling behaviors, and lessons learned},
  author={Ganguli, Deep and Lovitt, Liane and Kernion, Jackson and Askell, Amanda and Bai, Yuntao and Kadavath, Saurav and Mann, Ben and Perez, Ethan and Schiefer, Nicholas and Ndousse, Kamal and others},
  journal={arXiv preprint arXiv:2209.07858},
  year={2022}
}

@article{sun2023safety,
      title={Safety Assessment of Chinese Large Language Models}, 
      author={Hao Sun and Zhexin Zhang and Jiawen Deng and Jiale Cheng and Minlie Huang},
      journal={arXiv preprint arXiv:2304.10436},
      year={2023}
}

@inproceedings{sorensen2024value,
  title={Value kaleidoscope: Engaging ai with pluralistic human values, rights, and duties},
  author={Sorensen, Taylor and Jiang, Liwei and Hwang, Jena D and Levine, Sydney and Pyatkin, Valentina and West, Peter and Dziri, Nouha and Lu, Ximing and Rao, Kavel and Bhagavatula, Chandra and others},
  booktitle={Proceedings of the AAAI Conference on Artificial Intelligence},
  volume={38},
  number={18},
  pages={19937--19947},
  year={2024}
}

@article{tran2025multi,
  title={Multi-agent collaboration mechanisms: A survey of llms},
  author={Tran, Khanh-Tung and Dao, Dung and Nguyen, Minh-Duong and Pham, Quoc-Viet and O'Sullivan, Barry and Nguyen, Hoang D},
  journal={arXiv preprint arXiv:2501.06322},
  year={2025}
}

@article{luo2025large,
  title={Large language model agent: A survey on methodology, applications and challenges},
  author={Luo, Junyu and Zhang, Weizhi and Yuan, Ye and Zhao, Yusheng and Yang, Junwei and Gu, Yiyang and Wu, Bohan and Chen, Binqi and Qiao, Ziyue and Long, Qingqing and others},
  journal={arXiv preprint arXiv:2503.21460},
  year={2025}
}

@article{qwen25,
  title={Qwen2.5 technical report},
  author={Qwen Team},
  journal={arXiv preprint arXiv:2412.15115},
  year={2024}
}

@misc{NextAI,
    title = {The Second Half},
    author = {Shunyu Yao},
    year = {2025},
    url = {https://ysymyth.github.io/The-Second-Half/},
    note = {Accessed: 2025-05-13}
}

@article{liu2025scales,
  title={The Scales of Justitia: A Comprehensive Survey on Safety Evaluation of LLMs},
  author={Liu, Songyang and Li, Chaozhuo and Qiu, Jiameng and Zhang, Xi and Huang, Feiran and Zhang, Litian and Hei, Yiming and Yu, Philip S},
  journal={arXiv preprint arXiv:2506.11094},
  year={2025}
}

\end{document}